\documentclass[journal=jacsat,manuscript=article]{achemso}

\usepackage[T1]{fontenc}       
\usepackage{caption}
\usepackage{float}
\usepackage{geometry}
\usepackage{natbib}
\usepackage{setspace}
\usepackage{xkeyval}
\usepackage{amsmath}
\usepackage{booktabs} 
\usepackage{colortbl} 
\usepackage{xcolor} 
\usepackage{multirow}
\usepackage{adjustbox}
\usepackage{graphicx}
\usepackage{subfig}
\usepackage{multirow}
\usepackage{amssymb}
\usepackage{diagbox}
\usepackage{slashbox}

\author{Yashar Kiarashinejad}
\author{Sajjad Abdollahramezani}
\author{Ali Adibi}
\email{ali.adibi@ece.gatech.edu}
\affiliation{School of Electrical and Computer Engineering,	Georgia Institute of Technology, 778 Atlantic Drive NW, Atlanta, GA 30332, USA}

\title[\textsf{achesmo} demo]
{Deep learning approach based on dimensionality reduction for designing electromagnetic nanostructures}

\begin{document}


\begin{abstract}

In this paper, we demonstrate a computationally efficient new approach based on deep learning (DL) techniques for analysis, design, and optimization of electromagnetic (EM) nanostructures. We use the strong correlation among features of a generic EM problem to considerably reduce the dimensionality of the problem and thus, the computational complexity, without imposing considerable errors. By employing the dimensionality reduction concept using the more recently demonstrated autoencoder technique, we redefine the conventional many-to-one design problem in EM nanostructures into a one-to-one problem plus a much simpler many-to-one problem, which can be simply solved using an analytic formulation. This approach reduces the computational complexity in solving both the forward problem (i.e., analysis) and the inverse problem (i.e., design) by orders of magnitude compared to conventional approaches. In addition, it provides analytic formulations that, despite their complexity, can be used to obtain intuitive understanding of the physics and dynamics of EM wave interaction with nanostructures with minimal computation requirements. As a proof-of-concept, we applied such an efficacious method to design a new class of on-demand reconfigurable optical metasurfaces based on phase-change materials (PCM). We envision that the integration of such a DL-based technique with full-wave commercial software packages offers a powerful toolkit to facilitate the analysis, design, and optimization of the EM nanostructures as well as explaining, understanding, and predicting the observed responses in such structures. It will thus enable to solve complex design problems that could not be solved with existing techniques. In addition, this technique is not limited to the design of EM nanostructures; it can be easily extended to solve many different design problems in a wide range of disciplines as long as enough data for training of the incorporated neural networks is provided.

\end{abstract}

\section{1. Introduction}

The field of nanophotonics has been the subject of extensive expansion due to the unique capabilities of photonic nanostructures to control the propagation of EM waves. Owing to their constituent nanoscale features, which spectrally, spatially, and even temporally manipulate the optical state of the EM wave, nanophotonic devices extend all the functionalities realized by conventional optical devices in much smaller footprints. Combined with the advances in nanofabrication technologies, these nanostructures have been used to demonstrate devices with enormous potential for groundbreaking technologies addressing major challenges in state-of-the-art applications, such as optical communications \cite{melikyan2014high}, signal processing \cite{zhu2017plasmonic}, biosensing \cite{rodrigo2015mid}, energy harvesting \cite{liu2011taming}, and imaging \cite{huang2013three}, to name a few. As an example, newly-emerged metasurfaces (MSs)  \cite{yu2011light,kildishev2013planar,arbabi2015dielectric,khorasaninejad2016metalenses,jahani2016all,lin2014dielectric,taghinejad2018ultrafast,abdollahramezani2015analog,abdollahramezani2018reconfigurable}, two-dimensional planar structures comprising of densely arranged periodic/aperiodic arrays of well-engineered dielectric or plasmonic inclusions, offer profound control of the EM wave dynamics including amplitude, phase, polarization, and frequency in the subwavelength regime\cite{sun2012high,arbabi2017planar,decker2015high,chen2018broadband}. 

Despite extensive achievements in the fabrication and realization of photonic nanostructures, the efforts on the development of accurate and computationally efficient design and optimization approaches for these nanostructures are still at early stages\cite{molesky2018inverse}. With the fast progress in forming more complex nanostructures with several design parameters, the need for new design approaches that can keep pace with the computational requirements for analysis and understanding of all possible design options has become more imminent. In addition, realization of next-generation nanodevices with potentially new physics enabled through light-matter interaction at the nanoscale requires significant knowledge about the role of different design parameters in the functionality of a nanostructure. 

Traditional design and optimization approaches for EM nanostructures rely on either using analytical (or semi-analytical) modeling \cite{Piggott2017,Lu2013,Su2018,Frellsen2016,Piggott2014,Englund2005,molesky2018inverse} or brute-force analysis of the nanostructure through exhaustive search of the design parameter space \cite{Seidel1994}. The use of these approaches are limited to simple structures that could be either analytically modeled or completely studied by an exhaustive search technique with reasonable computation cost. To improve the computation efficiency of such design and optimization tools, evolutionary approaches (e.g., genetic algorithm \cite{Gondarenko2008,Hakansson2005} and particle swarm \cite{Ma2013}) rely on starting from a random initial guess and converging to the final optimum. While reducing the computation cost compared to brute-force approaches, such techniques are not guaranteed to converge to the global optimum of a problem (even by allocation of extensive computational resources). They are also limited to a single design problem (i.e., the simulations must be completely repeated when a small change in the nanostructure happens) and are computationally expensive for large-scale problems due to the significant amount of iterations to find the optimum design for a given device functionality.
\begin{figure}[tp]
	\centering
	\vspace*{-2cm}
	\includegraphics[trim=0cm 0cm 0cm 0cm,width=16.6cm,clip]{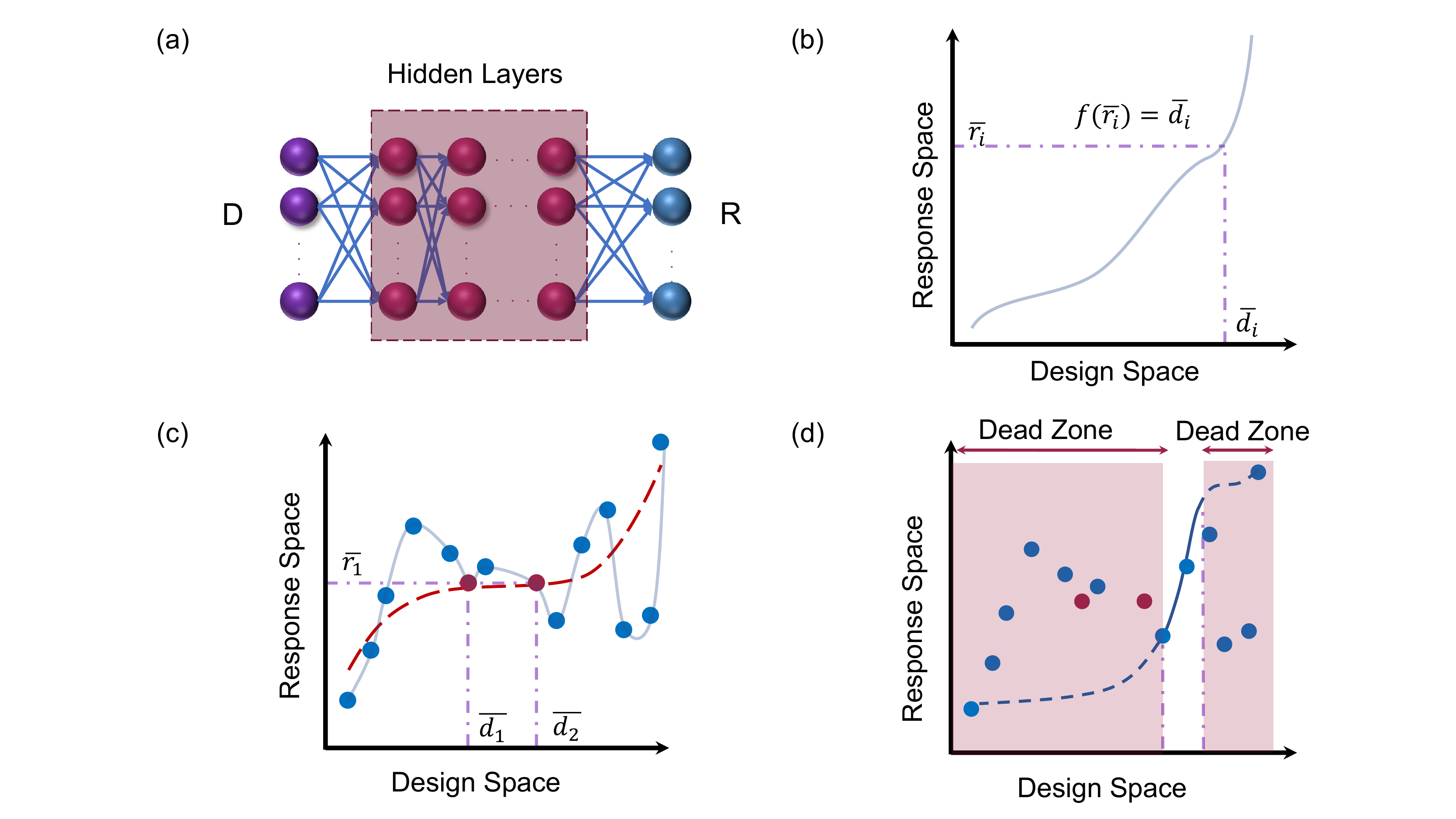}
	\caption{(a) A feed-forward NN for design and analysis of EM nanostructures; D and R represent design and response parameters, respectively. (b) Representation of a one-to-one design landscape (or manifold) as the simplest class of problems for solution with the NN in (a). (c) Representation of a general (non-one-to-one or many-to-one) design manifold. Red dots represent instances with same response features obtained with different sets of design parameters. The light-blue curve demonstrates the original design manifold while the dashed line shows the estimated one obtained with conventional methods for solving one-to-one problem (e.g., the NN in(a)). (d) Representation of the same deign manifold as in (c) with a solution obtained by just training the NN in (a) for some intrinsically one-to-one region (outside the dead-zones); the non-optimal extrapolated manifold for the dead-zones is highlighted by red color.}
	\label{fig:fig1}
\end{figure}

More recently, design and optimization approaches based on DL techniques have been proposed and implemented for the design of nanostructures \cite{Liu2018a,Peurifoy2018,Liu2018,Tahersima2018,Zhang2018a,Ma2018,Qu2018,Inampudi2018}. Different reported approaches to date primarily rely on training a neural network (NN)(see Fig. 1(a)) using the response of a set of devices (found by numerical simulations) and using the trained NN to solve the inverse design problem. Despite impressing progress in this area, the reported solutions mostly focus on solving simple problems with reasonably smooth optimization landscapes \cite{Peurifoy2018}that have a one-to-one mapping of design parameter space to the device response space (i.e., any given response can be obtained by only a single set of design parameters) as shown in Fig. 1(b), where a vector of device response ($\bar r_i$) is achieved by a unique vector of design parameters ($\bar d_i$). Unfortunately, most nanostructures of interest do not have this property. Figure 1(c) shows the optimization landscape of a more general problem in which the one-to-one relation between design parameters and output response does not exist. This can result in convergence issues for the NN used for optimization (i.e., finding design parameters for a given output response). Efforts on converting the problem to a one-to-one mapping by removing some training data sets (see Fig. 1(c)) \cite{Kabir2008} do not essentially help in solving the problem as most of the design space is not covered by these training datasets. Such approaches at most result in a NN that smooths out the optimization dataset (see Fig. 1(c)) without converging to the global optimum. Other proposed approaches (e.g., the use of tandem networks \cite{Liu2018a}) rely on first training a NN that relates the design space to the response space (i.e., for the forward problem), then cascading it as a pre-trained NN with another NN that relates the response space to the design space (i.e., the inverse problem), and finally training the resulting network (from the response space to the design space) to avoid the non-one-to-one relation. However, such techniques do not solve the main problem; they at best smooth out the optimization landscape as shown in Fig. 1(c). Another notable recent approach is based on using generative adversial networks (GANs) to solve the inverse design problem \cite{Liu2018}. This technique is built on training a network to solve the forward problem with zero error and use it to generate ground truth data in each iteration. Training such a forward-problem-solver network with zero error in a general design problem is a  major challenge and may require excessive computational resources. In the reported design problem, each desired output needs extensive computation (200,000 iterations to reach the convergence region for each structure)\cite{Liu2018}, which may reduce the value of using GANs if a perfect forward problem solver exists with comparable computation complexity (similar computations can be used to solve the design problem by exhaustive search using the perfect forward-problem solver). Despite impressing results, the reported GAN-based approach will be limited to simple design problems with non-complex nanostructure.  One can also think about breaking the NN for the design problem into two parts where one part is trained by limiting the design space to a smooth (one-to-one) region and the second part is trained by the data in the complete design space. The success of such techniques highly depends on the complexity of the problem and the selection of the design parameters in the one-to-one region (outside the dead-zones in Fig. 1(d)) to converge to acceptable answers. As a result, these approaches can be used to design simple structures, which can also be designed using alternative approaches. Finding a reliable approach to fundamentally address this non-uniqueness issue (without limiting the optimization landscape to the one-to-one region (or extrapolating from it, see Fig. 1(d)) is still a major challenge in using DL based approaches for the design of EM nanostructures. 

Another challenge in using DL techniques to design complex EM nanostructures is the large size of the response and design spaces resulting in the need to train a large NN. As an example, to study the spatial and spectral response of a MS with reasonable accuracy, the response space must constitute the sampled EM intensity in a two-dimensional space and in frequency with spatial and spectral resolutions smaller than the smallest spatial and spectral features of the output response, respectively. This typically results in thousands of data points in the response space and quickly rises as the structures with sharper spatial and spectral features are designed. Combined with ever-increasing number of design parameters in the nanostructures of recent interest, this results in a very large NN, which is difficult to be trained, even for the problems with one-to-one optimization landscapes.
 
In this paper, we demonstrate a new approach for designing complex EM nanostructures by addressing both the network-size issue and the non-uniqueness issue. Our approach is based on reducing the dimensionality of both the design space and the response space through training multi-layer NNs, called autoencoders \cite{Hinton2006}. Once the dimensionality of the problem is reduced, the problem converts into a one-to-one problem in the reduced spaces, which can be solved with considerably less computational complexity. In addition, by reducing the envisioned design parameters to few number of more complex design parameters (e.g., a nonlinear function of the weighted sum of the original design parameters), we can obtain valuable intuitive understanding of the roles of different design parameters in the response of the nanostructure. Such an efficacious approach paves the way for understanding, design, and optimization of complex EM nanostructures with far less computation complexity than the alternative approaches. In addition, a trade-off between the accepted error and the complexity (and time) of the simulations can be used to solve different problems with desired degrees of computation complexity or to obtain quick (approximate) information about the role of design parameters in the overall device performance. Dimensionality reduction (DR) is a powerful technique in machine learning that has been used to effectively solve problems in a wide range of applications including robotics \cite{Ciocarlie2007}, optical tomography \cite{Bhowmik2016}, face recognition \cite{Burger2000}, handwritten digit classification \cite{Hinton1997}, remote sensing\cite{efremenko2014optical}, medical science\cite{breger2017supervised}, genetics\cite{Kim2003}, and electronics \cite{Choi2017}.

To shows the applicability of our approach, we demonstrate its use for designing novel reconfigurable MS based on PCMs to form wideband amplitude modulation of near-infrared (near-IR) light. The rest of the paper is organized as follows. Section 2 describes our approach based on DR in detail. Section 3 shows the application of this approach for designing MS for the desired functionality along with the corresponding results. Section 4 describes the advantage of our approach in investigating the underlying physics . Section 5 is dedicated to the discussion of the obtained results and the unique features of the new approach. Final conclusions are made in Section 6.

\section{2. Dimensionality reduction of the design and response spaces in designing electromagnetic nanostructures}
\begin{figure}[t!]
	\centering
	\includegraphics[trim=0cm 0cm 0cm 0cm,width=12cm,clip]{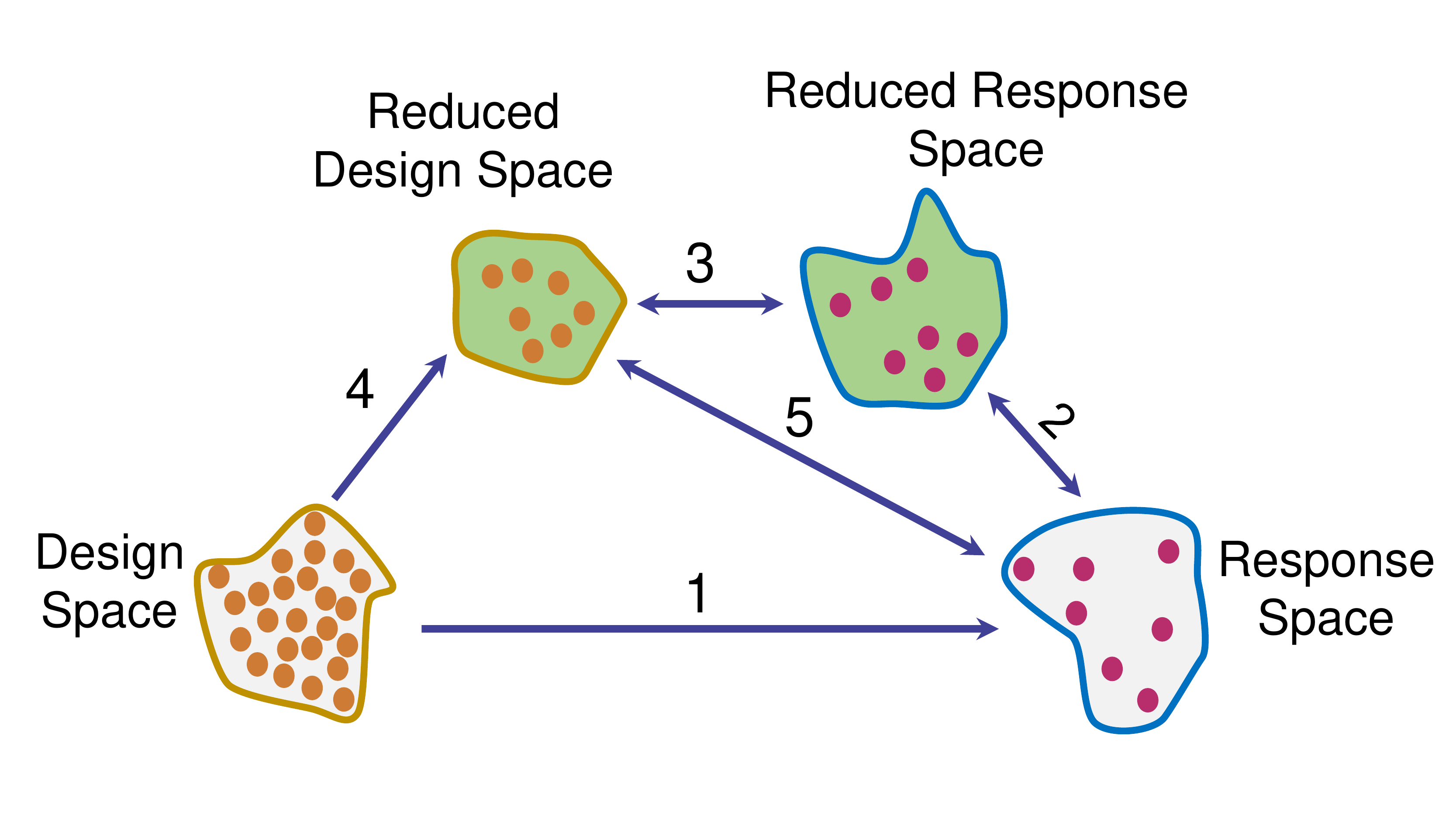}
	\caption{Application of the DR technique to the response and design spaces. In an optimal implementation, paths 1 and 4 are many-to-one while paths 2, 3, and 5 are one-to-one. The directions of arrows indicated paths that are easily achievable due to their one-on-one nature.}
	\label{fig:fig2}
\end{figure}

Figure 2 shows the schematic of the design approach based on DR of the design and response spaces assuming that the optimization landscape is non-unique (or many-to-one), i.e., more than one set of design parameters can result in the same response. The original forward problem is shown by path 1 in Fig. 2, where each point in the design space (that includes a vector of dimension D corresponding to a set of design parameters) correspond to a point in the response space (which includes a vector of dimension R) through a many-to-one relationship. A NN cannot be trained to inverse this relation as explained above. This is the main complication in the design and optimization problem. In our approach, we first use the DR technique to reduce the dimensionality of the response space as much as possible (i.e., reducing the size of the response vector $\bar r_i$ in Fig.1(b)) while keeping the same number of points in the response space (see path 2 in Fig. 2). This concept is schematically shown in Fig. 3 in which a three-dimensional manifold in the response space is reduced to a two-dimensional manifold, which includes the same number of points in the response space, but each point is represented by a smaller size vector. Each feature in the reduced response space is related to the features of the original response space through a well-defined nonlinear function. This is a one-to-one process.

In the next step, we reduce the dimensionality of the design space as much as possible (see path 4 in Fig. 2). In this process, the redundant nature of the design space is removed resulting in a one-to-one relation between the reduced design space and the reduced response space (see path 3 in Fig. 2). After training the relevant DR mechanisms in Fig. 2, the relation between the original response space and the reduced design space (Paths in Fig.2) will be one-to-one and thus, it can be simply inverted. Thus, our design problem will relate the desired response to the reduced design parameters (see path 5 in Fig. 2). The reduced design parameters are related to the original design parameters through a one-to-many relation that are analytically available through the training process. Thus, we can find several design options by converting the resulting optimum reduced design parameters to several sets of the original design parameters. At this stage, design constraints (e.g., fabrication imperfections, structure robustness, characterization limitations, etc.) can be taken into account to choose the final design parameters.

The heart of our approach is the effective implementation of the DR technique to maximally reduce the dimensionality of both design and response spaces, especially the former. Several DR techniques have been developed in machine learning to facilitate classification, data visualization, reduction of the computation cost, etc. Among different options, principal component analysis (PCA) \cite{Jolliffe2002}, kernel principal component analysis (KPCA) \cite{Sch1998}, Laplacian eigen map \cite{Belkin2003}, locally linear embedding \cite{Roweis2000}, and autoencoder \cite{Hinton2006} are the most effective techniques. Considering the features of these techniques, we believe that the autoencoder  is the most suitable approach for solving inverse problems in general and designing EM nanostructures in particular.

 \begin{figure}[t!]
	\centering
	\includegraphics[trim=0cm 0cm 0cm 0cm,width=8.3cm,clip]{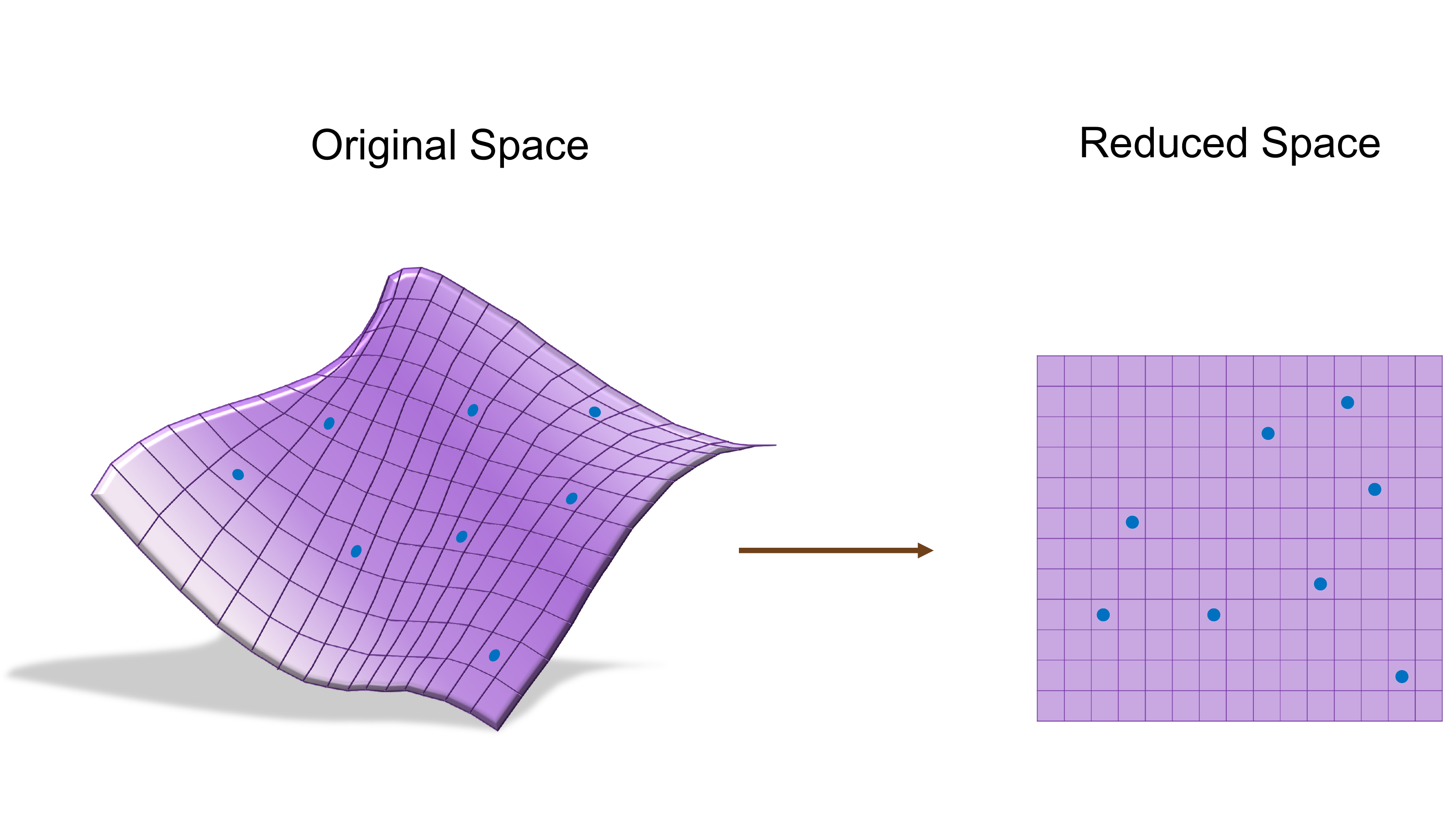}
	\caption{An example for the one-to one DR. Each dot represents a point in the original space that corresponds to a point (shown by a dot) in the lower-dimensional reduced space.}
	\label{fig:fig3}
\end{figure}
The general schematic of an autoencoder is shown in Fig. 4. Autoencoder is a multilayer NN that can encode the high-dimensional data into low-dimensional data (using the encoder part in Fig. 4) and use another NN (see the decoder part in Fig. 4) to decode and recover the high-dimensional data. In other words, the autoencoder in Fig. 4 is a feed-forward NN where the input layer and the output layer have the same structure and are connected to each other with one or more hidden layers. The number of neurons in the layer with minimum number of neurons represents the dimension of the low-dimensional data (i.e., the dimension of the DR space). This layer is known as the bottleneck of the autoencoder. This way, an autoencoder concentrates the data from a high-dimensional manifold in a given space around a low-dimensional manifold or a small set of such manifolds.The goal of an autoencoder is to map an original set of input data $\{x_1, x_2, \cdots, x_n\}$to a lower dimensional set of output data $\{s_1, s_2, \cdots, s_n\}$(at the bottleneck) in which $x_i$ and $s_i$ are vectors with size $k\times1$ and $m\times 1$, respectively($m<k$), and $s_i$ contains the essential information of $x_i$.

 To find the mapping from high-dimensional to low-dimensional data, the autoencoder in Fig. 4 should be trained with a sufficiently large training dataset. The training part of the autoencoder can be considered as an optimization problem where the algorithm minimizes a cost function. The cost function is a measurement of discrepancy between the output of the autoencoder and the input data. The mean-squared error (MSE) is used as the cost function of the autoencoder, and the error is minimized using the backpropagation method \cite{Rumelhart2013}. Assuming the output of the autoencoder structure in Fig. 6 for the input $x_i$ is represented by $\hat{x_i}$, the reconstruction MSE of the trained autoencoder is defined as:
 \begin{equation} 
 \label{eq: MSE}
 MSE=\frac{1}{\acute{n}} \sum_{i=1}^{\acute{n}} \|x_i-\hat{x_i}\|^2_2
 \end{equation}

 Where $\acute{n}$ represents the number of validation (or test) instances (not used for training) that are used to validate the trained autoencoder (but not used for training) . The number of layers and the topology of the NN is also found using an ad-hoc method (by trial and error). The training dataset for the design of EM nanostructures is obtained by using numerical simulation of the structure using a random set of input design parameters. 
\begin{figure}[t!]
	\centering
	\includegraphics[trim=0cm 0cm 0cm 0cm,width=8.3cm,clip]{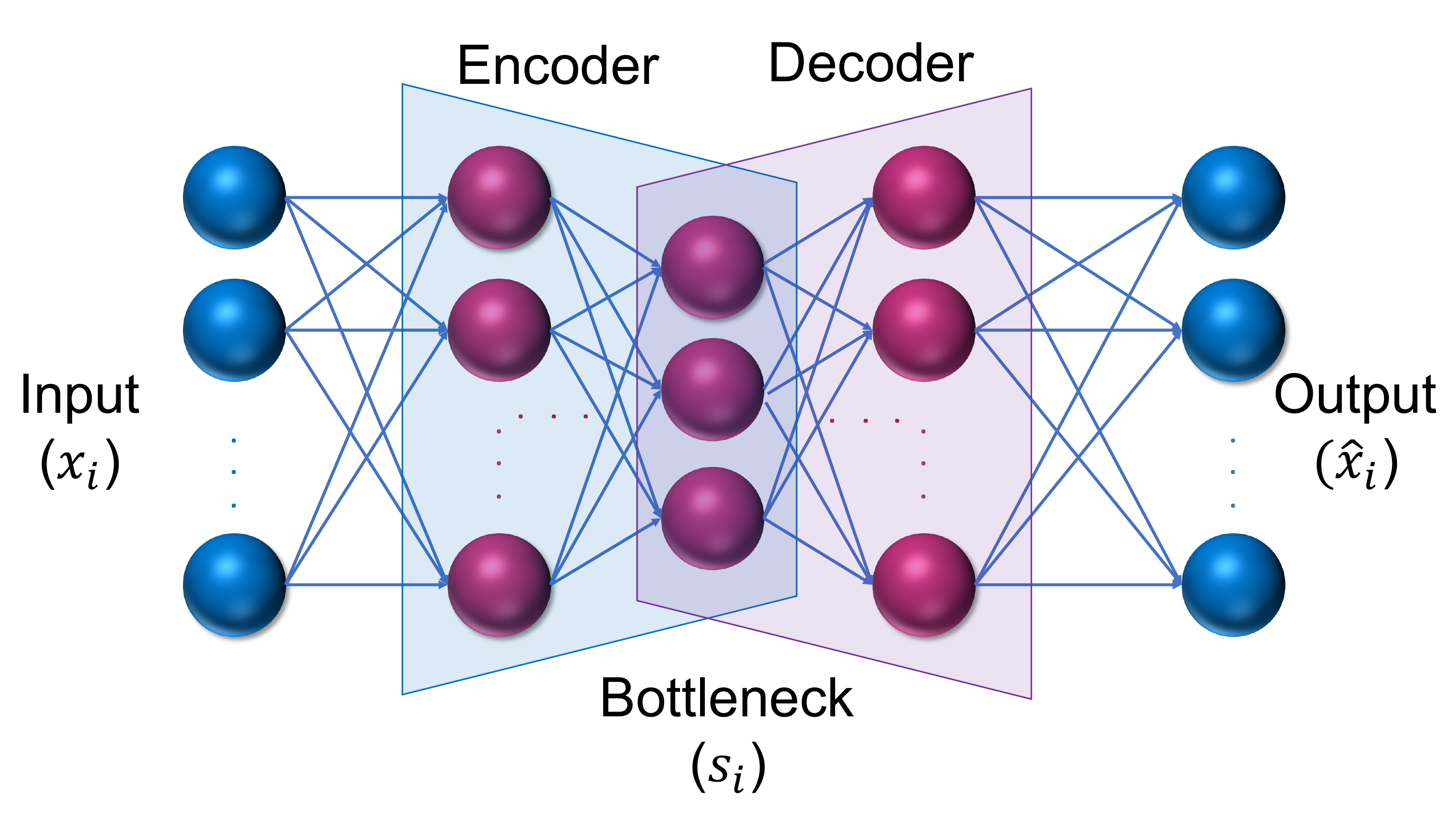}
	\caption{Schematic architecture of an autoencoder for the DR technique. The left half (i.e., encoder) reduces the dimensionality (the bottleneck layer corresponds to the reduced space) while the right half (i.e., decoder) brings the data back to the original space. The complete autoencoder is trained to minimize the MSE. }
	\label{fig:fig4}
\end{figure}

In the approach shown in Fig. 2, we first reduce the dimensionality of the response space by training an autoencoder (see Fig. 5(a)). In the next step, we form a pseudo-encoder that relates the original design space to the reduced response space as shown in Fig. 5(b). The reason for naming the structure in Fig. 5(b) a pseudo-encoder is the fact that its input and output are from different spaces (in contrast to a conventional autoencoder in Fig. 4). By training the pseudo-encoder in Fig. 5(b) to reach the minimum size of the bottleneck layer, we reach the reduced design space. Each parameter in this space is related to the original design parameters through a nonlinear function defined by the NN structure of the pseudo-encoder from the original design space to the reduced design space (or the bottleneck) in Fig. 5(b). The training approach is similar to that explained for a general autoencoder in Fig. 4 or Fig. 5(a). The pseudo-encoder in Fig. 5(b) corresponds to the paths 3 and 4 in Fig. 2, i.e., these two paths are trained together. 

\begin{figure}[t!]
	\centering
	\includegraphics[trim=0cm 0cm 0cm 0cm,width=16.6cm,clip]{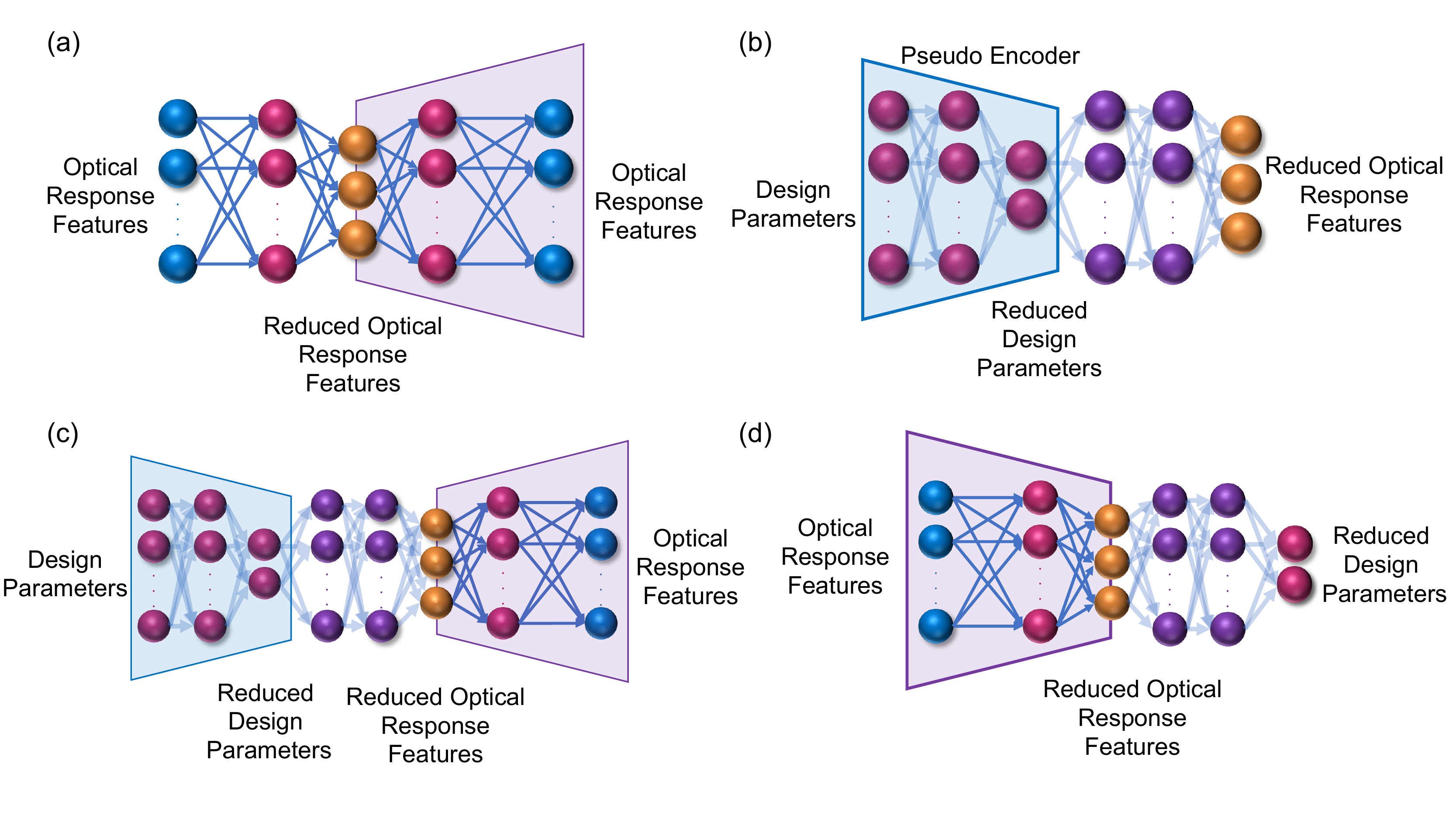}
	\caption{(a) Using an autoencoder to reduce the dimensionality of the response space (i.e., extract reduced optical response features from the original response features). (b) The pseudo-encoder architecture, which relates the original design space to the reduced response space while reducing the dimensionality of the design space (i.e., extracting the reduced design parameters). (c) A complete model for the forward problem formed by cascading the pseudo-encoder architecture in (b) with the decoder part of the autoencoder in (a). (d) The semi-inverse-problem model, which relates the original response space to the reduced design space as a one-to-one problem.}
	\label{fig:fig5}
\end{figure}

Once the DR of the two spaces are complete, we form a NN by cascading the pseudo-encoder in Fig. 5(b) with the pre-trained decoder part of Fig. 5(a) to form a completely trained NN for solving the forward problem as shown in Fig. 5(c). The resulting NN in Fig. 5(c) relates the original design parameters to the original response space using a unique set of analytic equations defined by different layers of the NN. While this analytic relation is complicated for a large-size network, it provides extremely valuable information about the roles of different design parameters in the response of the nanostructure with minimal computation complexity (technically by calculating the complex analytic formulas in a conventional environment like MATLAB). However, the goal of this paper is the design of EM nanostructures for which the inverse problem has to be solved.  For this purpose, we will use a two-step approach. In the first step, we find the inverse of the part of the NN in Fig. 5(c) that relates the reduced design space to the output space. The resulting inverse network is shown in Fig. 5(d). This is easily achievable as the relation between the reduced design space and the original response space is one-to-one (see path 5 in Fig. 2). The NN in Fig. 5(d) allows us to obtain the optimal reduced design parameters for any given desired response. This is the last part in our approach where the DL approaches can be used. The final step is to relate the reduced design parameters to the original design parameters (i.e., the inverse of path 4 in Fig. 2). This is a nonunique relation, i.e., it can provide several sets of design parameters from a given reduced set of design parameters. Fortunately, the encoder part of the pseudo-encoder in Fig. 5(b) relates the reduced design parameters to the original design parameters analytically (through the formulation of the underlying NN at different layers). We can use these equations to move layer-by-layer from the reduced design parameters to the original design parameters. In this backward process, we can reduce the number of possible solutions by imposing constraints such as fabrication limitations. This approach can provide many possible solutions for a design problem, which is expected due to the non-uniqueness of the problem. Note also that within this design problem, we can use the obtained knowledge about the role of the design parameters (using the forward solver in Fig. 5(c)) and the relation between the reduced design parameters and the original design parameters (using the encoder part of the pseudo-encode in Fig. 5(b)) to reduce the complexity in solving the design problem. In this paper, we use the analytic relation between the original and reduced design spaces to completely search the original design space to find the point(s) that correspond to the desired point in the reduced design space.

In addition to solving the non-uniqueness issue, the approach in Fig. 5 considerably reduces the computation cost by reducing the dimensionality of the two spaces. It is clear that the training of the pseudo-encoder that relates the design space to the reduced response space (see Fig. 5(b)) requires much less computation compared to training of a NN that relates the design space to the original (non-reduced) response space. Furthermore, the calculation of the inverse NN in Fig. 5(d) does not impose significant computation cost due to its one-to-one nature.

\section{3. Application to the design of hybrid reconfigurable plasmonic-photonic metasurfaces}
\begin{figure}[t!]
	\centering
	\includegraphics[trim=0cm 0cm 0cm 0cm,width=16.6cm,clip]{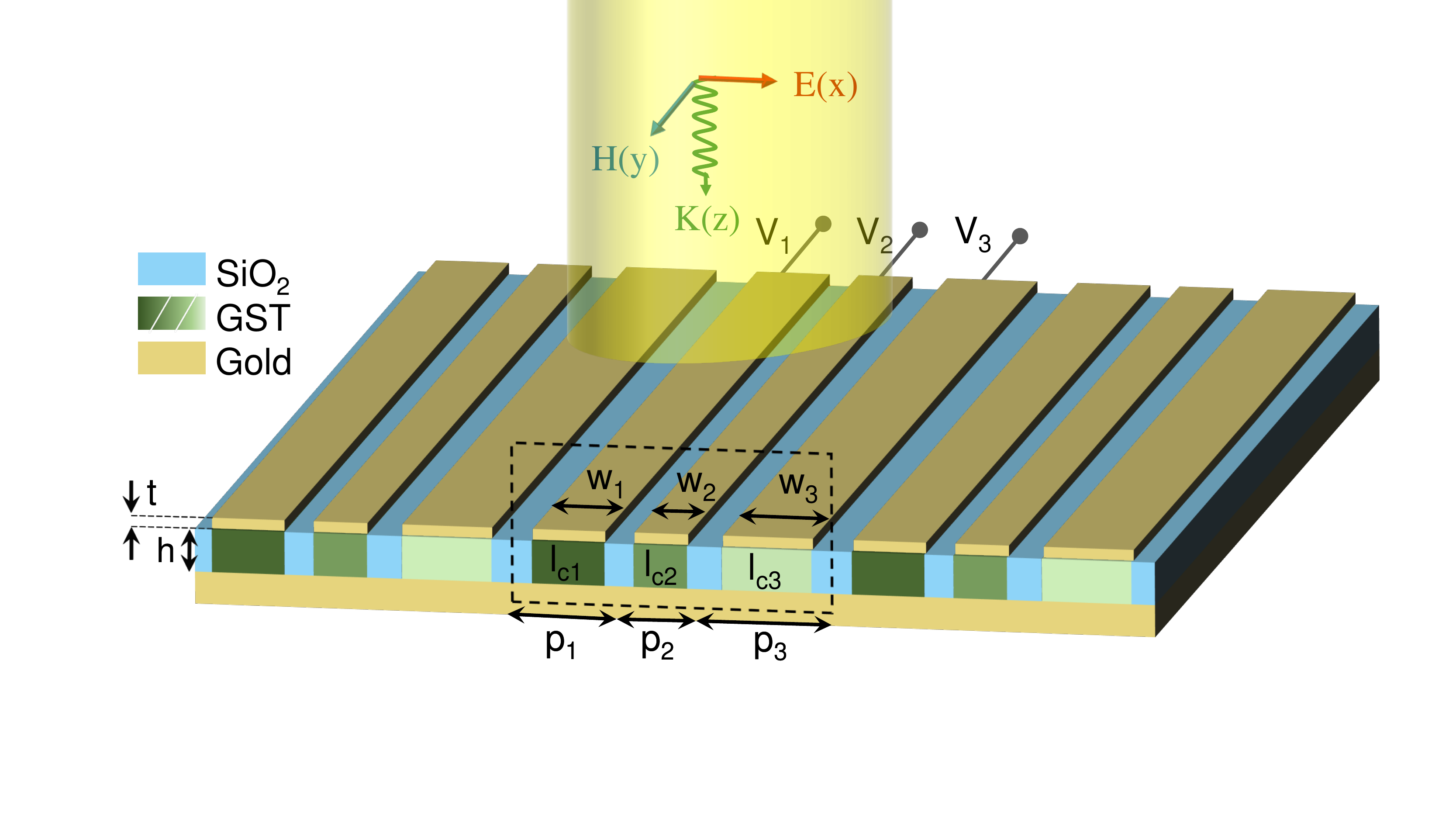}
	\caption{A MS with reconfigurable reflectivity formed by a periodic array of Au nanoribbons (thickness: t) on top of a thin layer (height: h) of GST on top of a SiO${_2}$ substrate. The unit cell of the structure is composed of three Au nanoribbons with different widths (w${_1}$, w${_2}$, and w${_3}$) and pitches (p${_1}$, p${_2}$, and p${_3}$, respectively). Other design parameters are the crystallization state of GST under the three nanoribbons (shown by l${_{c1}}$, l${_{c1}}$, and l${_{c3}}$, respectively) and the height of the GST layer(h). The phase of GST under each nanoribbon can be changed by applying a voltage (V${_1}$, V${_2}$, and V${_3}$, respectively). The incident light normally illuminates the MS, and the spatial and spectral profiles of the reflection from the structure is calculated as its response.}
	\label{fig:fig6}
\end{figure}
To show the applicability of the design approach, we consider a simple design problem for the implementation of a reconfigurable multifunctional MS enabling high performance optical modulation as shown in Fig. 6. The metasurface (MS) in Fig. 6 is composed of a periodic array of gold (Au) nanoribbons fabricated on top of a thin layer of germanium antimony telluride (Ge$_{2}$Sb$_{2}$Te$_{2}$ or in short GST), which is a non-volatile PCM whose index of refraction can be significantly modified (e.g., from 4.5 to 7 in the near-infrared region) \cite{wuttig2017phase}when it undergoes transition from the amorphous to the crystalline state in the near infrared regime or vice versa. In addition, using GST in intermediate states between amorphous and crystalline results in a wide range of tunability for its index of refraction. The GST layer is deposited on a thin film of Au as shown in Fig. 6. By applying electric signals to the Au nanoribbons, the state of the GST underneath that nanoribbon is controlled through resistive heating \cite{tuma2016stochastic}. In addition, by controlling the electric stimulus intermediate states (between amorphous and crystalline) can be obtained for GST \cite{rios2015integrated}. We limit the number of GST transition states to 11 (i.e., amorphous, crystalline, and 9 intermediate states)\cite{feldmann2017calculating}.  The supercell (limited to 3 different building blocks to prevent excitation of high diffraction orders) of the MS in Fig. 6 is composed of three Au nanoribbons with different widths (w${_1}$, w${_2}$, and w${_3}$) and three crystallization levels (l${_{c1}}$, l${_{c2}}$, l${_{c3}}$, corresponding to three indices of refraction, see Methods for more details) of GST underneath with the same height (h). The pitches of the 3 building blocks of the supercell are represented by p${_1}$, p${_2}$, and p${_3}$, respectively. As a result, the MS in this work has 10 design parameters (i.e., dimensionality of the design space is equal to 10) with different units (3 unit-less indices of refraction and 7 lengths with units of nanometers).

As a simple functionality, we are interested in amplitude modulation of the incident light at $\lambda=1600 nm$ with a considerable bandwidth around the central wavelength. The MS in Fig. 6 is illuminated with a plane wave of light with variable wavelengths in the desired range (from 1250 nm to 1850 nm). The polarization of the incident light is such that the electric field (i.e., E${_x}$) is perpendicular to the grating direction of the MS. The response of the system is the MS reflectance (calculated as the far-field reflection intensity divided by the intensity of the incident field and integrated over a surface area equal to one super-cell in the far-field). The resulting reflectance is sampled at 200 equally-spaced wavelengths in the 1250-1850 nm range. This results in a response space dimensionality of 200.  To obtain the data for training and validation of the DR autoencoders in Fig. 5, we simulate the structure in Fig. 6 with 4000 randomly generated instances (3600 for training and 400 for validation) formed by randomly selecting the design parameters in the acceptable variation ranges shown in the caption of Fig. 6. The simulations were performed using the finite element method (FEM) in the COMSOL Multiphysics environment (see Methods for details). 

In the next step, we use the training data to train a series of autoencoders with different numbers of hidden layers (ranging from 3 to 9) to study the DR of the response space from 200 to different values in the range of 1 to 20. For each autoencoder, MSE is calculated by finding the square of the norm of the difference between the reflectance vector obtained from the decoder part of the autoencoder and that obtained from the FEM simulations (also called ground truth data) for each one of the validation data. Note that each vector has 200 elements corresponding to the reflectance at 200 selected wavelengths in the 1250-1850 nm range.

Figure 7(a) shows the calculated MSE as a function of the dimensionality of the reduced response space. Figure 7(b) shows the comparison of the actual reflectance spectrum for the original data and the reconstructed data using the autoencoder for different dimensionalities of the reduced response space. It is clear from both Figs. 7(a) and 7(b) that the dimension of the response space can be reduced from 200 to 10 with negligible MSE (less than 10$^{-3}$). This is a clear advantage of our optimization technique. 
\begin{figure}[t!]
	
	\centering
	\includegraphics[trim=0cm 7cm 0cm 0cm,width=16.6cm,clip]{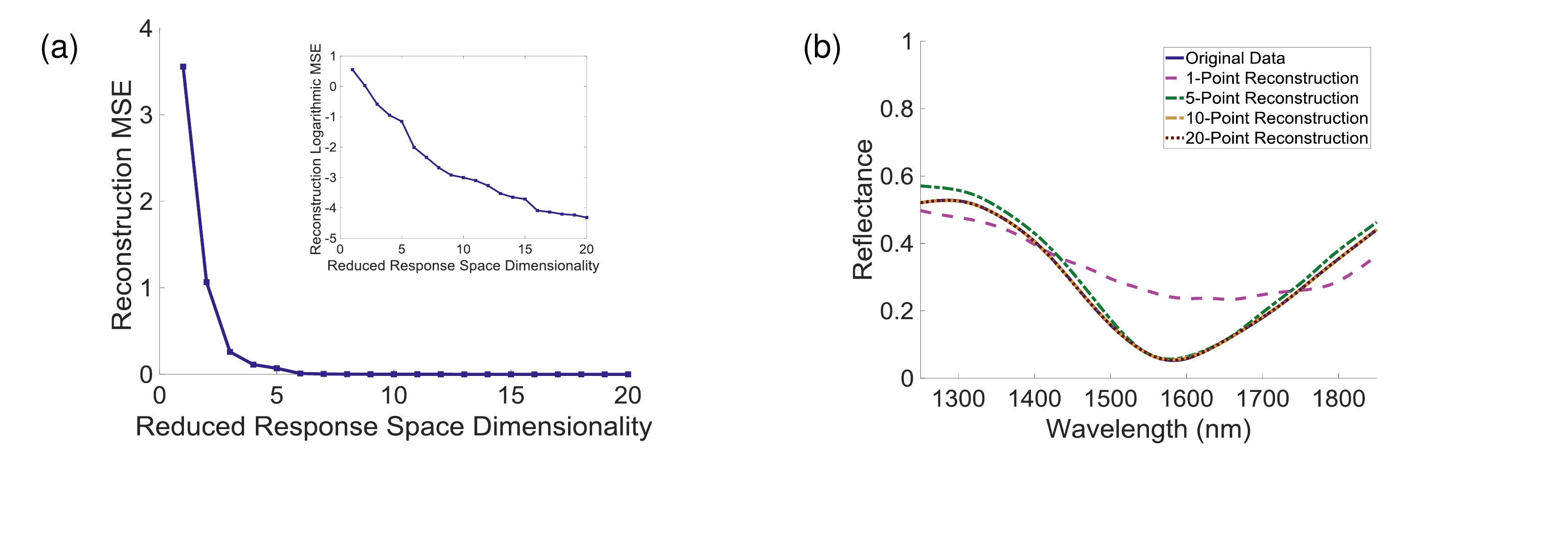}
	
	\caption{(a) MSE of the DR mechanism for the response space of the structure in Fig. 6 as a function of the dimensionality of the reduced response space. The inset shows the same data in the logarithmic scale. (b) The reconstructed response of the nanostructure in Fig. 6 after DR of the response space as a function of the dimensionality of the reduced response space.}
	\label{fig:fig7}
\end{figure}

Among different autoencoder architectures tested for the response space, the one with 5 layers (with the number of neurons in consecutive layers being 200-50-10-50-200) is selected based on its low MSE and computation costs to form the pseudo-encoder architecture in Fig. 5(b). We also choose 4 layers (10-20-15-x) for the encoder part of the pseudo-encoder in Fig. 5(b). For each set of values for the dimensions of the reduced response space and the reduced design space, we train the resulting pseudo-encoder using the 3600 training data, and we calculate the MSE by comparing the output of the pseudo-encoder with the actual output using the 400 validation data. The results for 4 different dimensionalities of the reduced response space are shown in Fig. 8(a). Figure 8(b) shows representative reflectance spectra for 3 different values of the dimensionality of the reduce design space. Using Fig. 8(a), it is clear that the dimension of the design space can be reduced from 10 to 5 without imposing much error.

Using Figs. 7 and 8, we choose the dimensionality of the reduced response space and the reduced design space to be 10 and 5, respectively (10-20-15-5-20-30-20-10-50-200). This considerably reduces the computation time as the dimension of the resulting problem is defined in a 5 $\times$ 10 rather than 10 $\times$ 200. Using these values, the final NN architecture for the analysis (or the solution of the forward problem) of the MS in Fig. 6 is formed according to Fig. 5(c). It is clear that the training of the pseudo-encoder that relates the design space to the reduced response space (see Fig. 5(b)) requires much less computation compared to training of a NN that relates the design space to the original (non-reduced) response space. 

\begin{figure}[t!]
	
	\centering
	\includegraphics[trim=0cm 7cm 0cm 0cm,width=16.6cm,clip]{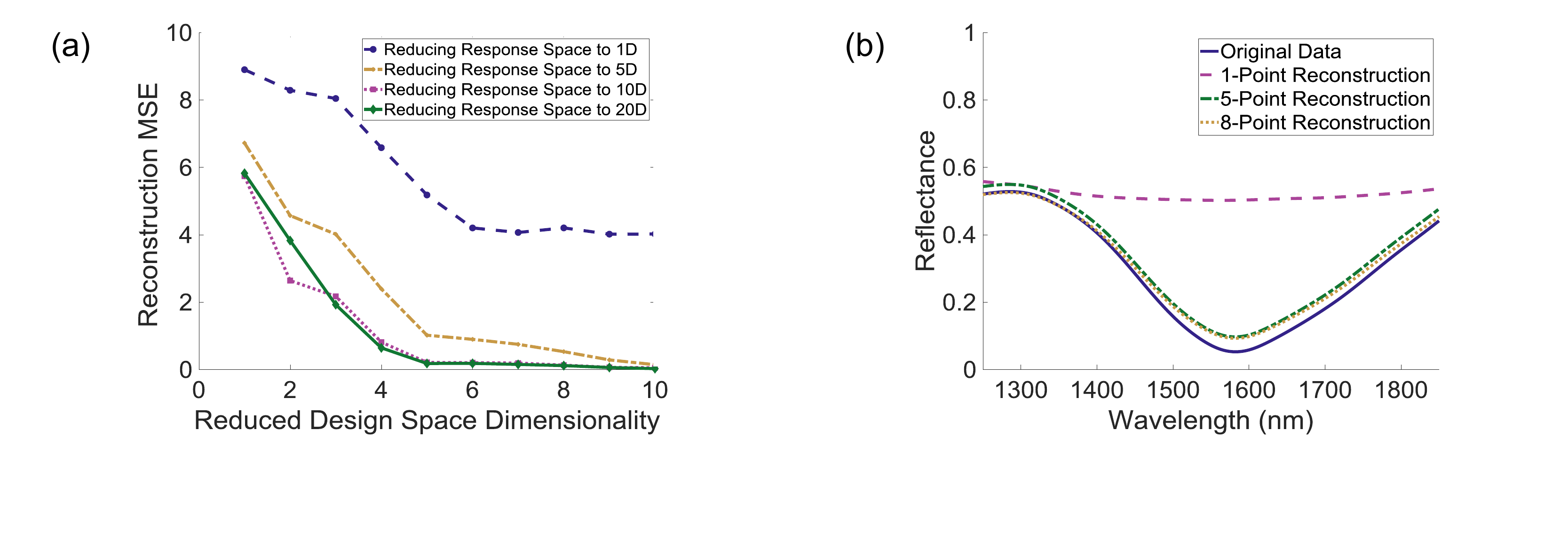}
	
	\caption{(a) Performance of the DR technique (i.e., the architecture in Fig. 5(c)) for the analysis of the response of the nanostructure in Fig. 6 (i.e., solution of the forward problem) in terms of MSE of reflectance for different dimensionalities of the reduced response and design spaces. (b) A typical spectral reflectivity response of the structure in Fig. 6 for different dimensions of the reduced design space with a fixed dimension (=10) of the reduced response space along with the original reflectivity response.}
	\label{fig:fig8}
\end{figure}
To form a platform for designing MSs with an arbitrary response, we first find the inverse of the network from the original response space to the reduced design space as shown in Fig. 5(d). This is not computationally extensive due to the one-to-one nature of the problem. For this purpose, the pre-trained encoder part of the DR algorithm for the response space (left side of Fig. 5(a)) is combined with a NN that connects the reduced response space to the reduced design space. This added NN is trained using the same 3600 training data to form the inverse network that relates the desired response to the reduced deign parameters. The resulting one-to-one trained platform (see Fig. 5(d)) results in finding the 5 reduced design parameters for the desired response. To find the 10 original design parameters, we solve the one-to-many problem through an analytical search approach using the encoder part of the pseudo-encoder for DR of the design space (first part of the platform in Fig. 5(b)). This encoder part relates the original design parameters analytically (through the NN formulation) to the reduced design parameters. Thus, the exhaustive search of the design space is not computationally extensive. We use MATLAB to perform this calculation (sweeping each parameter over 10 possible values) using the minimization of the MSE (defined by the integral of the square of the difference between the desired and the resulting light intensities over the operation bandwidth) as the optimization goal. 

Figure 9 shows the results for the design of prefect light absorber for operation in the 1500-1700 nm wavelength range using the MS in Fig. 6. The desired response is zero reflectivity over the entire operation bandwidth. The overall MSE for the response of the optimal structure in Fig. 9 is 0.0147 MSE. The reflectance for two other (non-optimal) designs with considerably different design parameters are also shown in Fig. 9. The set of design parameters along with the MSE for these three structures are listed in Table 1.

\begin{figure}[t!]
	\hspace*{0cm}
	\hspace*{3.5cm}
	\centering
	\includegraphics[trim=0cm 9cm 10cm 0cm,width=16.6cm,clip]{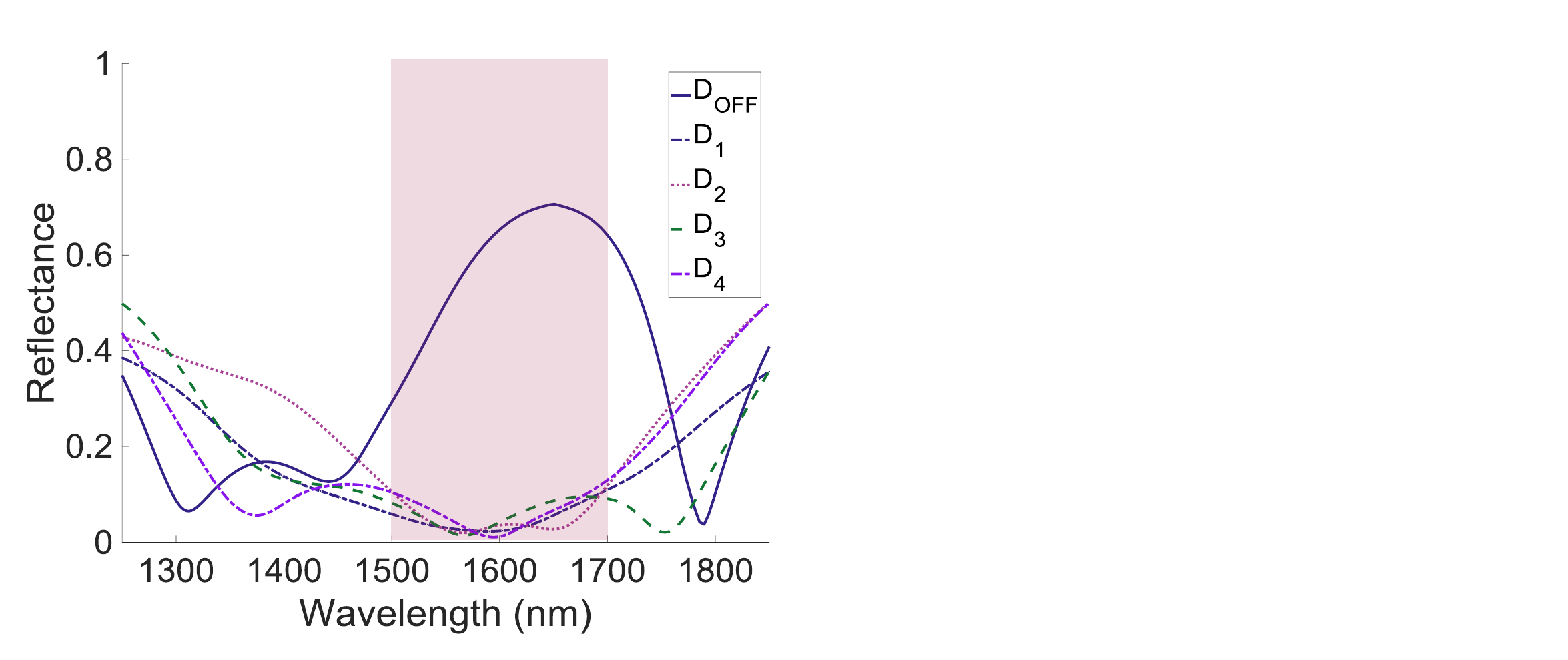}
	\caption{Responses of the optimal and three other reasonably good designed structures (Fig. 6, represented by dashed line) with the goal of achieving maximum absorption in the 1500-1700 nm wavelength region (shown by the shaded rectangle). The MSE and the values of the design parameters are shown in Table 1.The blue line show the reflectance spectrum of in the OFF-state (i.e., amorphous).}
	\label{fig:fig9}
\end{figure}

\section{4. Understanding the physics of light-matter interaction}
A main advantage of our approach is the possibility of investigating the underlying physics of the device operation and obtaining intuitive information about the roles of different design parameters on its response. To show this capability, we use our approach with a pseudo-encoder (10-4-10-50-20-10) to model the MS in Fig. 6. Figure 10(a) shows the resulting pseudo-encoder with the dimension of the reduced design space being 4 with green and red arrows representing positive and negative weights, respectively. Note that the DR of the design space is performed with only one encoder layer. Figure 10(b) shows the values of the weights for the mono-layer encoder. Each weight is multiplied by its corresponding design parameter to form the inputs to the node of the bottleneck layer. The larger the weight, the stronger the contribution of the corresponding design parameter will be. This strength is also shown in Fig. 10(a) by the thickness of the arrows that connect the nodes of the two layers. As shown in Fig 10(b), the height of the structure (h) plays an important role in changing the response compared to other design parameters as $h$ connects to all 4 nodes in the bottleneck layer with reasonably strong weights. Moreover, the crystallization levels l${_{c1}}$, l${_{c2}}$, and l${_{c3}}$ can only change one of the reduced response features as they mainly connect to only one node (the purple node) in the bottleneck layer. As a result, as long as the total input to that purple node is fixed, the response will stay the same regardless of how the values of l${_{c1}}$, l${_{c2}}$, and l${_{c3}}$ change. This conclusion is reached by assuming a small error in training the pseudo-encoder and neglecting the small weights (or arrows in Fig. 10(a)) that connect l${_{c1}}$, l${_{c2}}$, and l${_{c3}}$ to the nodes of the bottleneck layer. To test this conclusion, we vary l${_{c1}}$, l${_{c2}}$, and l${_{c3}}$ while keeping their weighted sum (according to the trained pseudo-encoder) and all other 7 design parameters for the MS in Fig. 6 constant, and we calculate the response of the MS using brute-force COMSOL simulations (no pseudo-encoder intervention). The results for two different weighted sums of l${_{c1}}$, l${_{c2}}$, and l${_{c3}}$ are shown in Fig. 10(c). Figure 10(c) clearly confirms our observation from the trained pseudo-encoder that l${_{c1}}$, l${_{c2}}$, and l${_{c3}}$ effectively act as one design parameter (through their weighted sum). Figure 10(d) shows the results of COMSOL simulations when the structure height ($h$) is changed while keeping all other 9 design parameters fixed. The large range of variation of the response in Fig. 10(d) clearly shows the importance of h as a design parameter. It is interesting to see from Fig. 10(d) that different responses for different values of h have low correlation while the responses for different values of the weighted sum of l${_{c1}}$, l${_{c2}}$, and l${_{c3}}$ (i.e., blues curves and red curves in Fig. 10(c)) show a similar trend with different locations of peaks and valleys. This suggests that the parameter $h$ can be used to obtain different classes of responses while the weighted sum of l${_{c1}}$, l${_{c2}}$, and l${_{c3}}$ can be used to finely tune a given class of response. The details of the design parameters for each case are shown in Table S2.

\begin{figure}[t!]
	\hspace*{0cm}
	\hspace*{0cm}
	\centering
	\includegraphics[trim=0cm 0cm 0cm 0cm,width=16.6cm,clip]{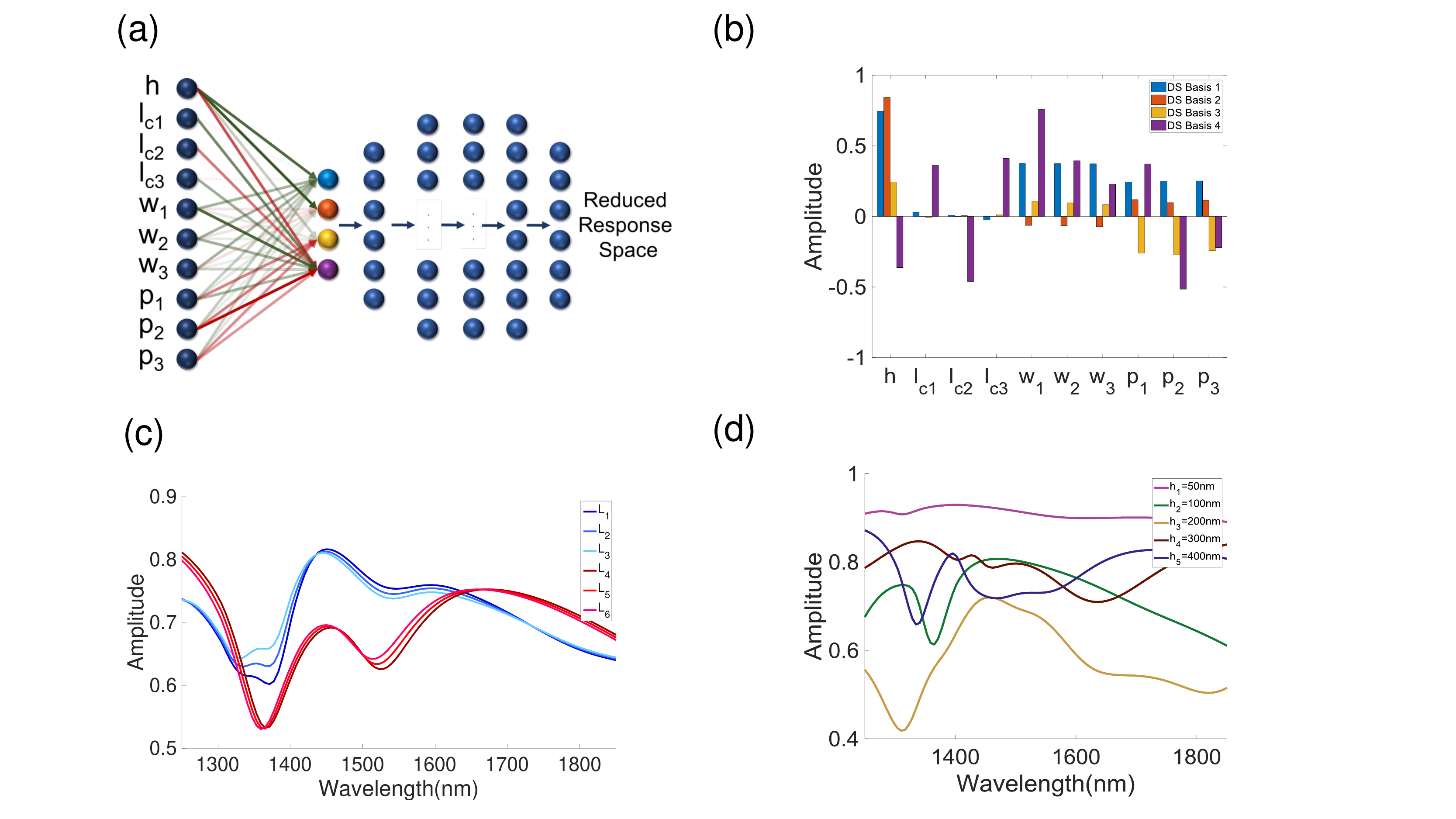}
	\caption{(a) The pseudo-encoder architecture trained for the problem in Fig. 6, which relates the original design space to the reduced response space while reducing the dimensionality of the design space Weights in a color map for the one-layer DR of design space. Color corresponds to the node. h has strengths to all 4 nodes, but l${_{c1}}$, l${_{c2}}$, and l${_{c3}}$ primarily connect to the purple. (b) Strength of the weights which are connected to different design parameters. (c) Red curves correspond to 3 different sets of l${_{c1}}$, l${_{c2}}$, and l${_{c3}}$ where their weighted sum for the purple node is the same. All other parameters are fixed. Blue: similar three curves but for a different weighted sum. (d) Variation of the response using COMSOL (no NN) where $h$ varies and all parameters are fixed. 
	}
	\label{fig:fig10}
\end{figure} 

The important observations about the role of different design parameters were obtained from our deep-learning approach without taking any information about the physics of the structure into account. Nevertheless, these observations agree with the physical intuition about the structure in Fig. 6. Each unit cell in this structure is composed of three plasmonic building blocks formed between the Au layer underneath and each Au nanoribbon on the top GST laye (see Fig. 6). Since the supermode of each building block is formed by coupling of the surface plasmon polaritons at the two Au layers, its properties strongly depend on the height of the GST layer (h), which directly controls the coupling strength\cite{maier2007plasmonics}. Thus, strong dependence of the  MS response on $h$ is expected. Figures 11 show the electric field patterns for the unit cell structure in Fig. 6 for two different values of h, confirming the strong dependence of the spatial mode profile on h. 
\begin{figure}[t!]
	\hspace*{0cm}
	\hspace*{0cm}
	\centering
	\includegraphics[trim=0cm 0cm 0cm 0cm,width=16.6cm,clip]{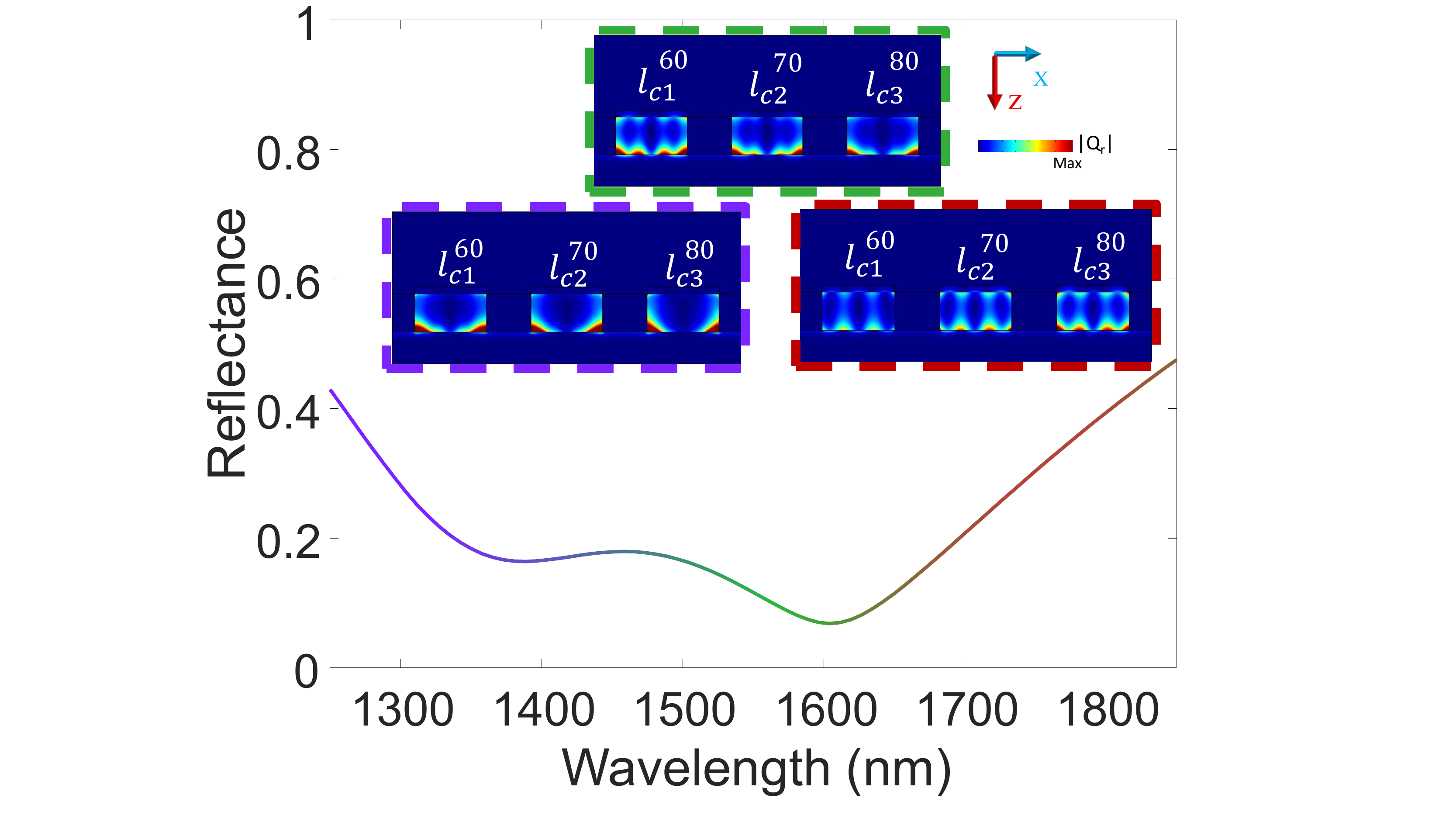}
	\caption{Simulated absorption spectra for the proposed broadband near-perfect absorber. Inset shows the resistive loss corresponding to the lower wavelengths (i.e., 1250 nm-1450 nm), intermediate wavelengths (i.e., 1450 nm-1650 nm), and higher wavelengths (i.e., 1650 nm-1850 nm) of the curve. $l_{c1}$, $l_{c2}$, and $l_{c3}$ represent $60\%$, $70\%$, and $80\%$ crystallization level of the GST layer, respectively.}
	\label{fig:fig11}
\end{figure} 

To consider the effect of variation of the crystallization fractions (l${_{c1}}$, l${_{c2}}$, and l${_{c3}}$), we note that the reflection response of the overall MS is essentially the sum of three responses defined by the three plasmonic resonators in each unit cell. By combining three wideband resonances with different resonance wavelengths, a wideband reflection response is obtained. Figure 11 shows the variation of the reflectance of the MS with frequency for a given set of l${_{c1}}$, l${_{c2}}$, and l${_{c3}}$ values (60\%, 70\%, and 80\%, respectively). The inset shows the field profiles of the three plasmonic resonators within each unit cell at different wavelength regimes.  

Due to pronounced light-matter interaction of the supermode with the GST layer at the higher wavelength (i.e., 1650-1850 nm), we expect that most of the resistive loss occurs in the building block with high crystallization level (i.e., l${_{c3}}$) accommodating more free charge carriers. This effect is clarified in the inset of Fig. 11 at higher wavelengths (red border) showing that a good portion of absorption takes place in the rightmost building block (i.e., l${_{c3}}$). Figure 11 also shows that the absorption loss in the middle wavelength window (i.e., 1450-1650 nm, shown by green) occurs mostly in the building block with lower crystallization level  (i.e., the leftmost building blocks with (i.e., l${_{c1}}$ and l${_{c2}}$). Finally, Fig. 11 shows similar contributions from the three building blocks at lower wavelengths (e.g., 1250-1450 nm). This is due to the fact that by increasing the level of crystallization in this regime, the optical constant of GST varies significantly leading to decrease in the light-matter interaction. This explains the collective role of l${_{c1}}$, l${_{c2}}$, and l${_{c3}}$ observed through training the pseudo-encoder. Note that obtaining this observation from the basic device properties was not as trivial as that of the role of $h$. 

While some of the conclusions about the role of design parameters in Fig. 6 obtained by training the pseudo-encoder could also be obtained by the underlying mode properties of the DL-based (e.g., by analyzing the modes of the plasmonic resonators), the ability of our approach in providing useful information about the physics of wave-matter interaction in non-trivial structures (e.g., nonlinear and dispersive metamaterials) will be extremely valuable. Indeed, by using this approach to find and understand new phenomena in such non-trivial structures, new ideas for forming novel device can be generated. This is a major advantage of our approach over all existing design approaches, especially those that rely on multiple brute-force simulations of the structure for different design parameters.

\section{5. Discussion}

Figure 9 and Table 1 clearly show the ability of our approach in designing MSs with considerably reduced computation complexity. Figure 9 obviously verifies the remarkable modulation depth (between ON- and OFF-state) of the optimized structure. They also show the importance of the understanding of the many-to-one nature of the design problem. Considering the many options for the original design parameters that correspond to a single set of reduced design parameters, we can easily enforce the fabrication restrictions and other design preferences in the last part of the design approach and find a set of design parameters that results in close-to-optimal response. It is important to note that the availability of the analytic relation between the original and reduced design spaces makes the brute-force optimization (e.g., using analytic search) computationally feasible even for a large number of design parameters. Nevertheless, more sophisticated constrained optimization techniques can be used to solve the last (i.e., many-to-one) part of the design problem with explicit inclusion of the fabrication and other design-related constraints. Such techniques are currently under investigation and will be the subject of future publications. 
\begin{table}[!b] 
	\centering 
	\begin{tabular}{l c c c c c c c c c c c c c} 
		\toprule 
		& \multicolumn{10}{c}{\textbf{Design Parameters}} \\ 
		\cmidrule (l){2-11} 
		\textbf{Design} & h & l${_{c1}}$& l${_{c2}}$& l${_{c3}}$& p${_1}$& p${_2}$& p${_3}$& w${_1}$& w${_2}$& w${_3}$ &MSE\\ 
		\midrule 
		$D_1$ & 190&0.5&	0.6	&0.7&	650&	650&	550&	350&	500&	200&	0.0147
		\\ 
		$D_2$ & 190	&0	&0.2	&0.8	&650	&650&	350&	450&	250	&250&	0.0149
		\\ 
		$D_3$ &190&	0.5&	0.1&	0.7&	650&	450&	450&	200&	350&	300&	0.0152
		\\ 
		$D_4$ &190	&0.3	&0.6&	0.8	&650&	550&	550&	250&	300&	450&	0.0172
		\\ 
	
		\midrule 
		
	\end{tabular}
	\caption{The design parameters and the resulting MSE for the optimal design and three good designs for the structure in Fig. 6 to achieve maximum absorption in the 1500-1700 nm wavelength region.} 
	\label{tab:template} 
\end{table}

A unique feature of our DR-based approach is the computation simplicity while appreciating the many-to-one nature of the problem. By not considering the latter explicitly, several other existing NN-based techniques are technically limited to only smooth-enough problems, or they require  apriori assumptions to limit the search for the optimal solution in to a given region in the design space, where the relation to the response space is one-to-one (as discussed in Section 1). Nevertheless, by reducing the dimensionality of the problem, our approach requires less computation than any other alternative. For example, in the design problem studied here, we reduced a 10 $\times$ 200 dimensional problem to a 5 $\times$ 10 one. 

Compared to brute-force optimization approaches (e.g., exhaustive search), our technique requires far less computation. For example, by assuming only 10 possible values for the 7 analog variables (i.e., h, w${_1}$, w${_2}$, and w${_3}$, p${_1}$, p${_2}$, and p${_3}$) and 11 values for the discrete ones (i.e., l${_{c1}}$, l${_{c2}}$, l${_{c3}}$) in the design problem in Fig. 6, the exhaustive search algorithm requires the complete EM simulation of the structure for more than 10$^{10}$ times, which is essentially intractable. However, our optimization requires only 4000 EM simulations along with the training process that requires far less computations. Indeed, the entire training of the forward and inverse parts of the platform in Fig. 5 for the design problem in Fig. 6 (results shown in Fig. 9) took less than 3 hours using a simple personal computer with a 3.4 GHz core i7-6700 CPU and 8 GB of random access memory (RAM). 

The main issue that can happen in the non-optimal use of the DR platform is the non-uniqueness (or non-one-to-one) relation between the original response space and the reduced design space in the platform shown in Fig. 5(d). This can happen by not carefully considering different numbers for the reduced dimensions in the DR algorithms. Nevertheless, the DR algorithm considerably reduces the possibilities of the many-to-one problem by limiting the dimensionalities of the design and response spaces. Thus, the resulting structure in Fig. 5(d) can provide a close-to-optimal design even for the cases that the resulting architecture in Fig. 5(d) is not absolutely one-to-one. A rigorous mathematical study of the conditions for the dimensionality of the reduced spaces from machine-learning point of view can provide more solid guidelines in selecting the dimensionality of the reduced spaces. However, such a rigorous mathematical study is outside the scope of this paper.

\section{6. Conclusion}

		We demonstrated here a new DL-based approach for the design of EM nanostructures with a wide range of design possibilities. We showed that by reducing the dimensionality of the response and design spaces using an autoencoder and a psuedo-encoder, we can convert the initial many-to-one problem into a one-to-one (or in the worst case, close to one-to-one) problem plus a simple one-to-many problem that can be solved using brute-force analytical formulas. The resulting approach considerably reduces the computational complexity of both the forward problem and the inverse (or design) problem. In addition, it allows for the inclusion of the design restrictions (e.g., fabrication limitations) without adding computational complexities. It also provides valuable information about the roles of design parameters in the response of the EM structure, which can potentially enable novel phenomena and devices. Finally, this technique can be extended to solve many different optimization problems in a wide range of disciplines as long as enough data for training the incorporated NNs are provided. 
\setcounter{figure}{0}
\setcounter{table}{0}

\section{7. Methods}
The full-wave EM simulations were carried out using the finite element method (FEM) enabled by linking the commercial software package COMSOL Multiphysics 5.3 (wave optics module)  with MATLAB to expedite the design, optimization, and analysis processes. Floquet periodic and perfectly-matched layer boundary conditions were used along transverse x-axis and the z-axis in Fig. 6, respectively. The structure was assumed infinite along the y-direction. A linearly polarized planewave of light, excited the MS in the wavelength range of 1250-1850 nm. The refractive index (n) and absorption coefficient (k) data for amorphous and crystalline GST, Au, and SiO${_2 }$ were obtained from the literature \cite{shportko2008resonant,huang2016gate}. The computation domain was meshed using triangular elements with a maximum size of $\lambda_0$/(10 n) in SiO${_2}$ and GST , and of $\lambda_0$/50 in Au with $\lambda_0$ being the free-space wavelength. The effective dielectric constant associated with the intermediate states of GST were approximated via the effective medium theory. Among different options, Lorentz-Lorenz formula more accurately describes effective permittivity $\epsilon_{eff}(\lambda_0)$ as\cite{chu2016active}:
\begin{equation} 
\makeatletter 
\makeatother
\frac{\epsilon_{eff}(\lambda_0)-1}{\epsilon_{eff}(\lambda_0)+2}={l_c}\times\frac{\epsilon_{c}(\lambda_0)-1}{\epsilon_{c}(\lambda_0)+2}+({l_c}-1)\times\frac{\epsilon_{a}(\lambda_0)-1}{\epsilon_{a}(\lambda_0)+2},
\label{Lorentz-Lorenz}
\end{equation}
where $\epsilon_{c}(\lambda_0)$ and $\epsilon_{a}(\lambda_0)$ are the permitivitties of the crystalline and amorphous GST, respectively, and ${l_{c}}$, ranging from 0 (amorphous) to 1 (crystalline), is the crystallization fraction of GST. 
\section{8. Acknowledgements }
The authors thank Ali A. Eftekhar for helpful discussion.
\section{9. Author contributions}
YK and SA contributed equally to this work. The initial idea was developed by YK and SA, and its implementation was discussed by all authors. YK performed the training optimization of the autoencoder and the pseudo-encoder, and SA developed the simulation results for training and validation. SA proposed the initial idea for the electromagnetic nanostructure for light absorption, which was discussed in further details by all authors. AA managed the project. All authors participated in writing the manuscript. 
\section{10. Competing interests}
The authors declare no competing financial interest.
\newpage


\newpage

\bibliography{DRoct2018}

\end{document}